\documentclass[11pt, a4paper]{article}
\usepackage{cite}
\usepackage{subfig}
\usepackage[pdftex]{graphicx}
\graphicspath{{./Figures/}}
\DeclareGraphicsExtensions{.pdf,.png}

\usepackage[cmex10]{amsmath}
\usepackage{amsfonts}
\usepackage{authblk}
\begin{document}
\title{Momo: Monocular Motion Estimation on Manifolds}

\author[1]{Johannes Graeter}
\author[1]{Tobias Strauss}
\author[1]{Martin Lauer}
\affil[1]{Institute of Measurement and Control (MRT) \\
Karlsruhe Institute of Technology (KIT)\\
Email: johannes.graeter@kit.edu}


%Telephone: (800) 555--1212, Fax: (888) 555--1212}


\maketitle

% % As a general rule, do not put math, special symbols or citations
% % in the abstract
\begin{abstract}
Knowledge about the location of a vehicle is indispensable for autonomous driving.
In order to apply global localisation methods, a pose prior must be known which can be obtained from visual odometry. The quality and robustness of that prior determine the success of localisation.
\\
Momo is a monocular frame-to-frame motion estimation methodology providing a high quality visual odometry for that purpose. By taking into account the motion model of the vehicle, reliability and accuracy of the pose prior are significantly improved. We show that especially in low-structure environments Momo outperforms the state of the art.
Moreover, the method is designed so that multiple cameras with or without overlap can be integrated.
The evaluation on the KITTI-dataset and on a proper multi-camera dataset shows that even with only 100--300 feature matches the prior is estimated with high accuracy and in real-time. 
\end{abstract}

% % For peer review papers, you can put extra information on the cover
% % page as needed:
% % \ifCLASSOPTIONpeerreview
% % \begin{center} \bfseries EDICS Category: 3-BBND \end{center}
% % \fi
% %
% % For peerreview papers, this IEEEtran command inserts a page break and
% % creates the second title. It will be ignored for other modes.
% \IEEEpeerreviewmaketitle

% % inline sections
\section{A short story on monocular visual odometry}
\label{sec:a_shortstory_on_monocular_visual_odometry}

Visual odometry has been successfully applied for more than 20 years.
Especially the work of Hartley and Zissermann in the late 1990s builds the basis for modern visual odometry algorithms~\cite{hartley2003multiple}.
They introduced a new normalization method for the then already well known 8-Point-Algorithm, turning it into the standard frame-to-frame motion estimation algorithm.\\
For a calibrated camera the 8-Point-Algorithm is overdetermined. Therefore Nister et al.~\cite{nister2004efficient} proposed the 5-point-algorithm, which reduces the motion parameter space by an Eigenvalue decomposition.
However, for robots with non-holonomous motion patterns such as vehicles, the problem is still overdetermined.
There have been various attempts to adapt the problem to special motion patterns (Hee et al.~\cite{hee2013motion}, Scaramuzza et al.~\cite{scaramuzza2011}), however none of them replaced the 5-point-algorithm as standard frame-to-frame motion estimation algorithm.
\\
If a sequence of frames is used, the standard algorithm for estimating the motion of vehicles is Simultaneous Localisation and Mapping (SLAM).
By building a map of the environment, temporal information can be added to the problem in an effective way.
Since map and motion have to be estimated simultaneously, the amount of parameters for a full bundle adjustment is very large and therefore time consuming. Possessing a good frame-to-frame motion prior for the full bundle adjustment is hence crucial for real-time performance.
\\
\begin{figure}[!t]
\centering
\includegraphics[width=\columnwidth]{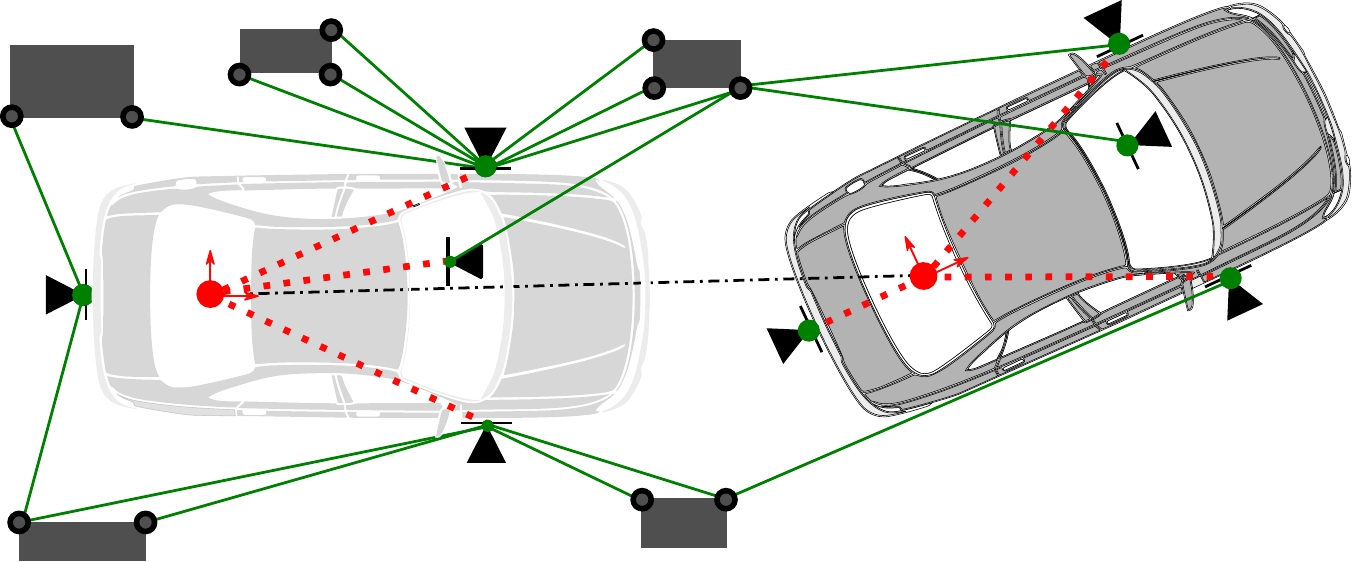}
\caption{Diagram illustrating Momo's problem formulation. Assuming a motion model in the motion center of the vehicle (red dots), an error metric is evaluated including all cameras (green dots).
Using this general formulation, the required number of features for correct frame-to-frame motion estimation can be reduced to 100--300.}%
\label{fig:motivation}
\end{figure}

In real-life environments, the main challenge is outlier handling, as shown by recent advances in stereo vision. 
Observing the 10 best stereo vision algorithms on the challenging KITTI dataset~\cite{geiger2013vision}\footnote{accessed on 13th of March 2017}, it is striking that all of them propose new methods for outlier rejection (Buczko et al.~\cite{buczko2016flow}, Cvivsic et al.~\cite{cvivsic2015stereo}), while not using bundle adjustment.
Impressively, without using temporal inference, they can obtain results with less than $1\%$ translation error, which demonstrates how accurate such algorithms can become if correct feature matches are chosen.
Outlier rejection for monocular systems is more challenging since no depth information is available.
In implementations such as libviso~\cite{geiger2011stereoscan} or the opencv library~\cite{bradski2008learning} a RANSAC algorithm is used for outlier rejection.
Thereby, the main assumption is that most visible features belong to the static scene. 
If big objects occlude the static scene this assumption is violated.
In the current state of the art, the focus of the work on monocular visual odometry is mostly on effective ways of bundling rather than on the careful selection of outliers during prior estimation -- outliers are typically filtered by heuristics, such as their distance to the map.
However, if the proportion of outliers is high, the quality of the motion prior becomes very important in terms of time consumption and accuracy.
\\ \\
The goal of Momo is the improvement of monocular visual odometry by more careful outlier selection and more accurate estimation of the pose prior.
In order to increase robustness against rough weather conditions or occlusion, the methodology is designed to be capable of estimating the motion even if only few feature matches could be established.
Furthermore multi-camera setups are supported, as shown in fig.~\ref{fig:motivation}. All available information is integrated by Momo into one single optimization problem, resulting in a reliable and accurate motion prior.
\\ \\
We publish the code and supplementary material such as a video showing the system in action on GitHub~(\emph{https://github.com/johannes-graeter/momo.git}).
\begin{figure*}[!t]
\centering
\includegraphics[width=0.9\textwidth]{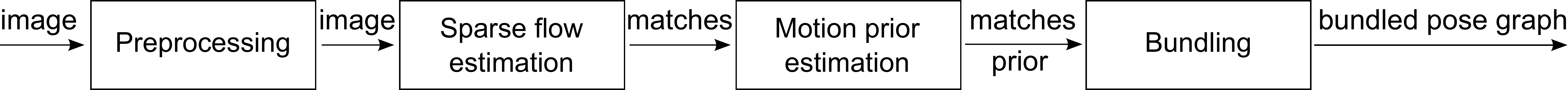}
\caption{Pipeline of the monocular visual odometry estimation procedure.}
\label{fig:pipeline}
\end{figure*}
\section{General problem formulation}
\label{subsec:general_problem_formulation}

In this section the approach proposed in this work is introduced and compared to existing methods. The contribution of this work concerns the prior estimation step, as shown in the visual odometry pipeline, fig.~\ref{fig:pipeline}.
Careful selection and tuning of feature matching methods are essential for a working system and will be explained in section~\ref{sec:results}.
For bundling, various excellent frameworks exist such as g2o~\cite{kummerle2011g} or gtsam~\cite{dellaert2012factor}.
\\
The input of the prior estimation block is a set of matched features $X$ between two frames of size $N$ that is called $x_{i,\tau}\in X, i\in [1\dots N], \tau\in [\tau_0,\tau_1]$.
This set can also be interpreted as sparse optical flow.
\\
Our goal is the extraction of the motion $M^{\tau_0}_{\tau_1}$ between these two consecutive frames containing six degrees of freedom.
Therefore, the solution of the optimization problem 
\begin{equation}
\label{equ:min_problem}
    M^{\tau_0}_{\tau_1}=\underset{M}{\mathrm{argmin}}(\mathcal{E}(X,M))
\end{equation}
is sought, where $\mathcal{E}(X,M)$ depicts the energy potential to be minimized.
\\
\subsection{Relation to the 8-point- and 5-point-algorithm}
\label{subsec:connection}
The 8-point- and 5-point-algorithm are linear solutions to equation~\ref{equ:min_problem}.
In these methods, the energy potential $\mathcal{E}(X,M)$ is the summed epipolar error, explained in section~\ref{subsubsec:formulating_the_potential_function}.
However, this error metric is non-linear. Therefore, a non-normalized, linear variant of the epipolar error is used $\mathcal{E}(X,M)=\overset{N}{\underset{i}{\sum}}x_{i,\tau_1}^T F(M) x_{i,\tau_0}$, with the fundamental matrix $F$ (see Hartley and Zisserman~\cite{hartley2003multiple}). Normalization is either applied on the input measurements, as done by Hartley et al.~\cite{hartley1997defense} or directly applied on the fundamental matrix, as proposed by Torr et al.~\cite{torr2004invariant}. 
\\
This is a valid solution for an outlier-free environment, since the problem is linear and therefore can be solved efficiently.
However, its main disadvantage is that these solutions are very susceptible to outliers.
In the state of the art, sampling based outlier rejection models such as RANSAC~\cite{fischler1981random} or LMEDS~\cite{rousseeuw2005robust} are wrapped around the problem in order to find the most plausible solution. Therefore, many hypotheses must be tested that violate the motion patterns of realistic systems.
\\
Enforcing motion models on the linear approach of Hartley~\cite{hartley2003multiple} is complicated and limited to simple motion patterns. 
For example in order to reduce the degrees of freedom from 8 to 5, a sophisticated Eigenvalue analysis is necessary as shown by Nister et al.~\cite{nister2004efficient}.

\subsection{Advantages of the non-linearised system}
As explained in section~\ref{subsec:connection}, the linear formulation of problem (\ref{equ:min_problem}) makes modelling non-holonomous movement by motion models difficult and outlier rejection more expensive.
\\
Therefore, a different approach is proposed herein, dropping the linearisation.
This results in the following advantages:
\begin{enumerate}
\item Implicit robustification of the problem by an M-estimator.
\item Optimization on manifolds of the motion space using a motion model for the vehicle.
\item Generalization of the problem to calibrated multi-camera systems without overlapping field of view.
\item Adaptation to general camera models.
\item Implicit scale estimation in curves.
\end{enumerate}

\section{Methodology}
\label{sec:methodology}
\subsection{Formulation of the potential function}
\label{subsubsec:formulating_the_potential_function}
In this section a reconstruction-free error metric for the potential function is formulated.
Popular choices for this error metric take advantage of the epipolar geometry defined by $M^{\tau_0}_{\tau_1}$ and a point correspondence $x_{i,\tau}$ between two images.
In this section a short overview of common error metrics concerning the epipolar geometry is given. For more detail we refer to Hartley and Zisserman~\cite{hartley2003multiple}.
In this work, the focus is on the following error metrics:
\begin{enumerate}
\item The geometric distance of $x_{i,\tau_1}$ to its corresponding epipolar line, called $GeoLine$.
\item The angle between the line of sight of $x_{i,\tau_1}$ and the epipolar plane, called $AnglePlane$.
\end{enumerate}

Note that $GeoLine$ is evaluated in the image domain, whereas $AnglePlane$ is evaluated in Euclidean space.
As a result $GeoLine$ is only usable for pinhole camera models, but $AnglePlane$ can be used for any camera model, including highly non-linear camera models.
\\
The distance of an image point to its corresponding epipolar line is defined as 

\begin{equation}
d(x_{i,\tau_1},F x_{i,\tau_0})=\frac{x_{i,\tau_1}^T F x_{i,\tau_0} }{\sqrt{(F x_{i,\tau_0})^2_1+(F x_{i,\tau_0})^2_2}} \text{,}
\end{equation}
where the denominator is used for normalisation.
$(\cdot)_i$ denotes the i-th row of the vector.
$F$ is the fundamental matrix defined as $F=K^{-T} E K^{-1}$.
Hereby, the camera intrinsics $K$ have to be known by camera calibration.
The essential matrix $E$ is fully defined by $M^{\tau_0}_{\tau_1}=\left[R|t\right]$ with $E(M)=[t]_{\times} R$, with the skew-symmetric matrix $[\cdot]_{\times}$. 
The metric  $d(x_{i,\tau_1},F x_{i,\tau_0})$ is not symmetric. Therefore, the so called geometric distance is more commonly used:
\begin{equation}
GeoLine=\overset{N}{\underset{i=1}{\sum}} d(x_{i,\tau_1},F x_{i,\tau_0})^2+d(x_{i,\tau_0},F^T x_{i,\tau_1})^2 \text{.}
\end{equation}
\\   
However, $GeoLine$ can only account for pinhole camera models.
To generalize the potential function, a non-linear camera model is considered, for which the lines of sight for each pixel in the image are known.
\begin{equation}
AnglePlane=\overset{N}{\underset{i=1}{\sum}} \frac{(\hat{x}_{i,\tau_1}^T E \hat{x}_{i,\tau_0})^2 }{\|E \hat{x}_{i,\tau_0}\|^2_2} \text{,}
\end{equation}
where $\hat{x}_{i,\tau}$ denotes the line of sight corresponding to $x_{i,\tau}$.
Note that for pinhole camera systems the line of sight can be calculated by  $\hat{x}_{i,\tau}=\frac{K^{-1}x_{i,\tau}}{\|K^{-1}x_{i,\tau}\|_2}$.
$AnglePlane$ is therefore the generalization of $GeoLine$ to non-linear, single-view-point camera models.

\subsection{Establishing a robust potential function}
\label{subsubsec:robustifying_the_potential_function}
A great advantage of using non-linear estimation for frame-to-frame motion estimation is the possibility to use robust loss functions in order to reduce the influence of outliers on the potential function, thus turning the problem into an M-estimator.
The goal is to reduce the influence of outliers, which is essential for finding the correct estimation.
On that account, a robust loss function $\rho(x)$ is wrapped around the energy potential $\mathcal{E}$.
Since the growth of $\rho(x)$ becomes small with increasing $x$, outliers are weighted down.
The potential function becomes therefore
\begin{equation}
    \mathcal{E}_{\mathrm{robust}}(X,M)=\rho(\mathcal{E}(X,M)) \text{.}
\end{equation}
Popular loss function choices are Tukey or Huber loss.
In the proposed system Cauchy loss is used. It neglects the influence of outliers, since $\underset{x \rightarrow \infty}{\mathrm{lim}}\frac {\mathrm{d}\rho(x)_{cauchy}}{\mathrm{d}x} = 0$, but is not down weighting as aggressively as Tukey loss. Therefore the use of Cauchy loss helps to avoid local minima. 

A big advantage compared to sampling-based outlier rejection schemes is that no random samples are needed and a motion prior can be considered.
Using the previous estimation as a prior significantly increases efficiency, since the motion transition is smooth in online scenarios.
In addition to that, only plausible solutions are tested.
As a result, the assumption can be dropped that the biggest amount of matches must belong to the static scene.

\subsection{Optimization on manifolds of the motion space}
\label{subsubsec:optimization_on_manifolds_of_the_motion_space}
\begin{figure}[!t]
\centering
\includegraphics[width=0.8\columnwidth]{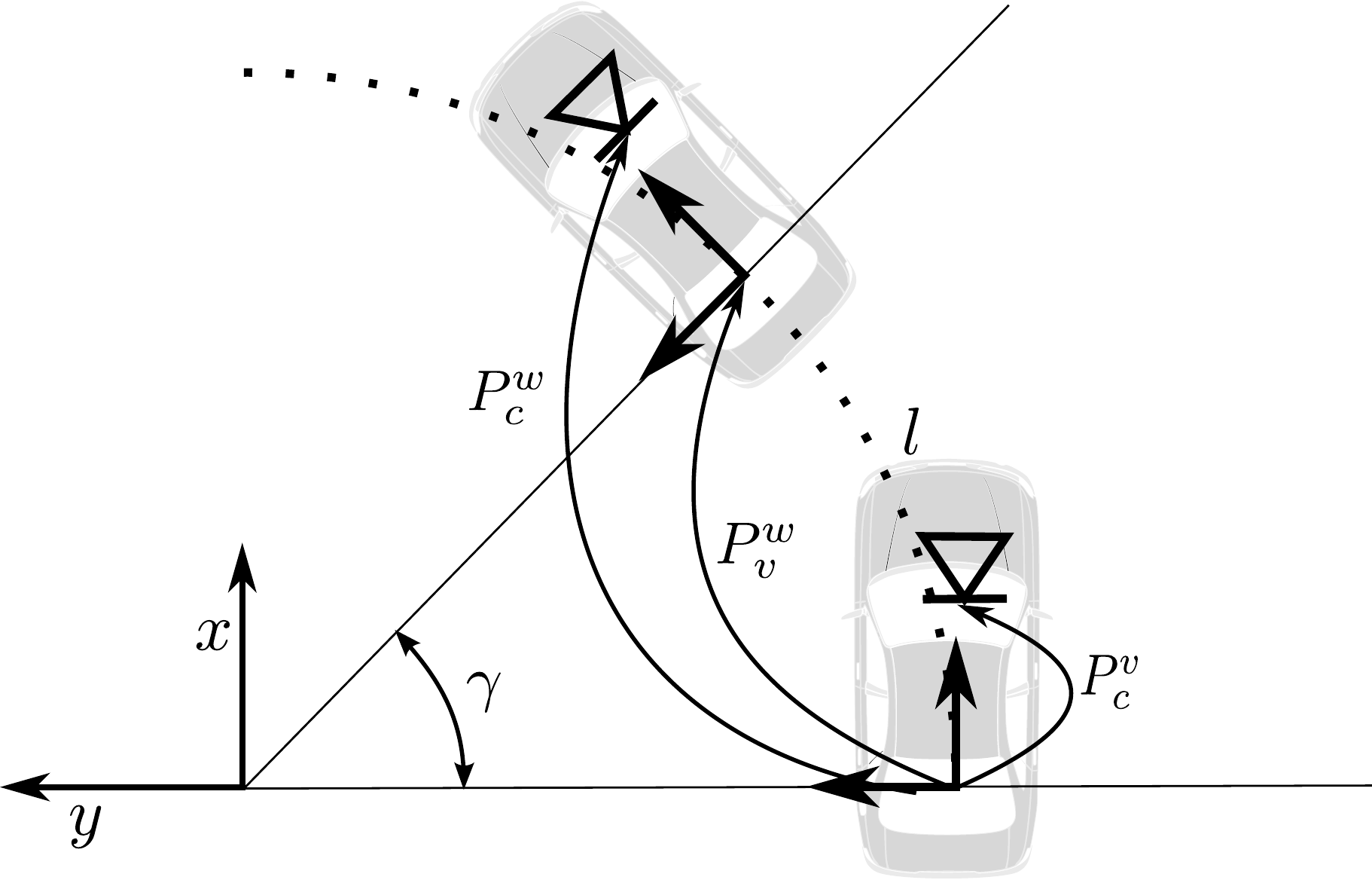}
\caption{Sketch of the frame-to-frame movement on a circle with a non-centred camera. $x$ and $y$ are global coordinates, $\gamma$ is the change of yaw angle, 
$l$ is the travelled arc length on the circle. Moreover the following transformations are defined: $P^w_v$, from the motion center at $\tau_0$ to the motion center at $\tau_1$; $P^w_c$, from the motion center at $\tau_0$ to the camera at $\tau_1$; $P^v_c$ the extrinsic calibration of the camera. }
\label{fig:sketch_scale}
\end{figure}

In a real world scenario the camera is mounted on a vehicle, which has non-holonomous motion characteristics.
Therefore, only a subspace of the full 6 degrees of freedom is used, the motion manifold.
For linear approaches such as the 8- and 5-point methods, great effort has to be done to reduce the motion space. Introducing complex models is non-trivial.\\
Momo was specifically designed to serve as prior estimation on a broad range of systems -- we constructed it so that any motion model can easily be integrated.
In this work, we model the motion of the autonomous vehicle with the well-known single-track-model.
Neglecting slip, this motion model describes a planar movement on a circle as shown in fig.~\ref{fig:sketch_scale}.

With the radius of the circle $r=\frac{l}{\gamma}$, the motion $P^w_v$ between the motion centres can be formulated as
\begin{equation}
\label{equ:motion_model}
P^w_v = \left ( \begin{array}{rrr|r} \cos(\gamma)&-\sin(\gamma)&0&\sin(\gamma)r \\ \sin(\gamma)&\cos(\gamma)&0&(1-\cos(\gamma))r \\ 0&0&1&0 \end{array} \right ) \text{.}
\end{equation}
Furthermore, the 2d model can be enhanced by pitch and roll angles thus resulting in a 3d model.
\\
Equation~\ref{equ:motion_model} is an example how motion models of any complexity can be applied for Momo, since only the mapping from the manifold to the full 5d motion space must be known.
\\
In general, the point from which the motion is origins is not the mounting position of the camera.
In case of an autonomous car the center of motion is usually the middle of the rear axis, whereas cameras need to be mounted externally.
With the general problem formulation, the extrinsic calibration, i.e. the transform from the motion center to the cameras, can be trivially taken into account by spatial transformation $P_{camera}={P^v_c}^{-1}{P^w_v}{P^v_c}$.
This formulation enables the implementation of various motion models and their integration into the potential function in order to consider a-priori knowledge.

\subsection{Using multiple cameras}
\label{subsubsec:using_multiple_cameras}
Section~\ref{subsubsec:optimization_on_manifolds_of_the_motion_space} describes how the extrinsic calibration of the camera to the center of motion can be included into the potential function.
Using this methodology, the problem can be expressed from the motion center to any point in space.
For usage on a multi-camera system with known extrinsic calibration, the problem formulation is trivial - in order to include several cameras into the problem, the excited motion is propagated from the motion center to each camera.
The minimzation problem of the total error therefore becomes
\begin{equation}
\label{equ:multiple_cameras}
\underset{M}{\mathrm{argmin}} \underset{j}{\overset{N}{\sum}} \mathcal{E}(X_j,P^{-1}_j M P_j) \text{,}
\end{equation} 
with the number of cameras $N$, matches $X_j$ and extrinsic calibration $P_j$ for each camera respectively. 
Using several cameras counteracts many of the problems that monocular perception has: 
\begin{itemize}
\item The surroundings can be perceived from many viewpoints, thus increasing the variety of measurements.
\item A camera that is occluded by rain drops or dirt can be detected if most of the other cameras are still working.
\item If the cameras are mounted on different sides of the vehicle, there is at least one camera that is not blinded by sunlight.
\end{itemize}

\section{Results}
\label{sec:results}
\subsection{Selection of the error metric}
In order to choose the most suitable cost function, we simulated sparse optical flow.
Since $AnglePlane$ is the generalisation of $GeoLine$ both show almost identical error landscapes, with a well defined minimum and a convex landscape.
Additionally, we evaluated convergence speed, where $AnglePlane$ shows fastest convergence.
The plots and more detailed information can be found on GitHub~(\emph{https://github.com/johannes-graeter/momo.git}).\\
Both error metrics, $GeoLine$ and $AnglePlane$, are suitable choices for the potential function.
In order to enable general camera models and obtain fast convergence, we chose $AnglePlane$.

\subsection{Evaluation on KITTI}

\begin{figure}[!t]
\centering
\subfloat[KITTI sequence 00,  Momo (dashed blue), 5-point (dotted orange)]{\includegraphics[width=\columnwidth]{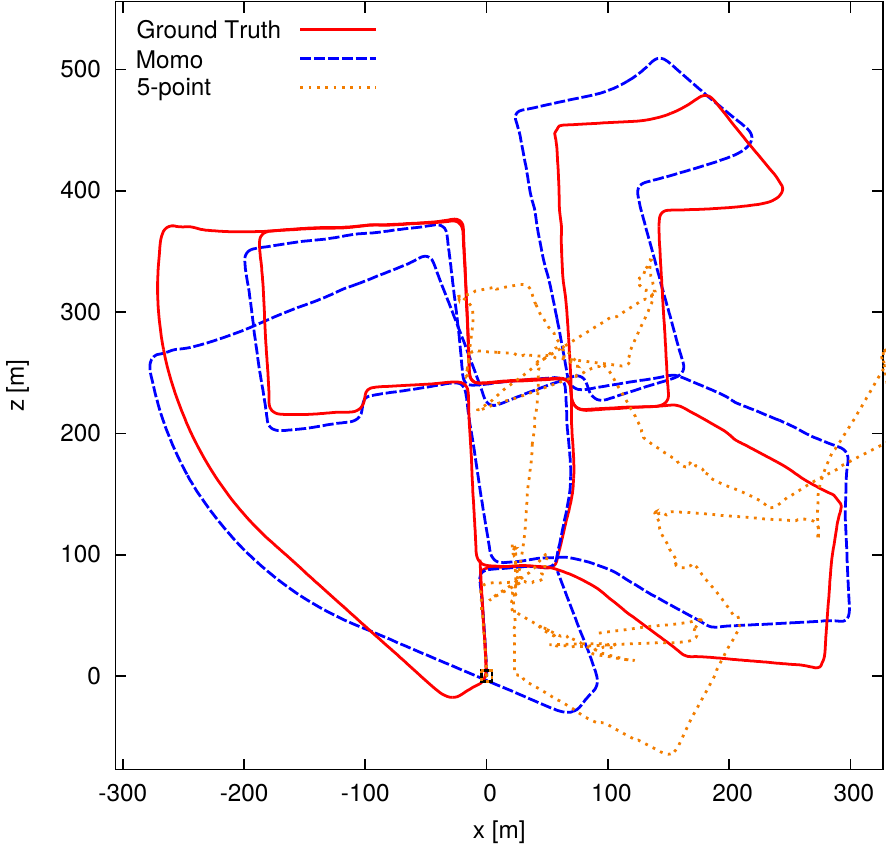}%
\label{fig:kitti_paths:00}}
\hfil
\subfloat[KITTI sequence 10,  Momo (dashed blue), 5-point (dotted orange)]{\includegraphics[width=\columnwidth]{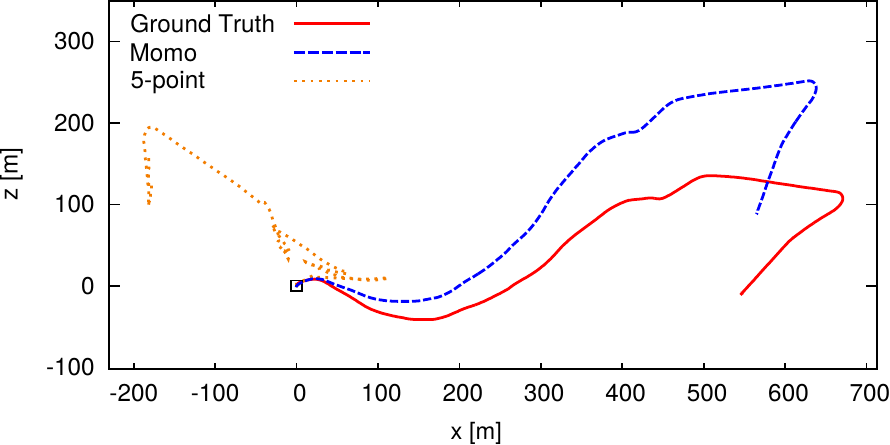}%
\label{fig:kitti_paths:10}}
\caption{Two examples of estimated trajectories from the KITTI dataset shown as topview. Since the method is designed as a prior estimator, it is evaluated frame-to-frame without drift reduction through temporal inference.
Scale is taken from groundtruth.
The feature matcher is tuned so that only 100--300 feature matches per image are available.
While the 5-point-algorithm (dotted orange) cannot estimate the path correctly, Momo (dashed blue) gives a very good frame-to-frame visual odometry.}
\label{fig:kitti_paths}
\end{figure}

\begin{figure}[!t]
\centering
\subfloat[Rotation error over length, Momo (blue), 5-point (orange)]{\includegraphics[width=0.95\columnwidth]{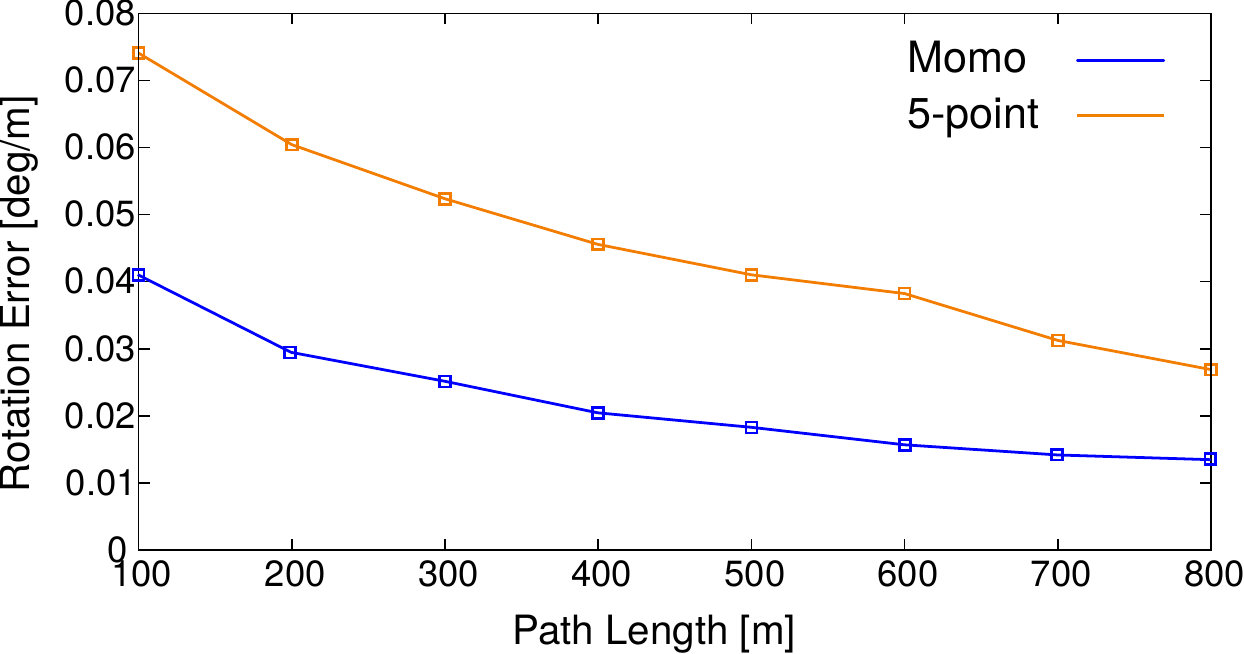}%
\label{fig:rotation_error:length}}
\hfil
\subfloat[Rotation error over speed, Momo (blue), 5-point (orange)]{\includegraphics[width=0.95\columnwidth]{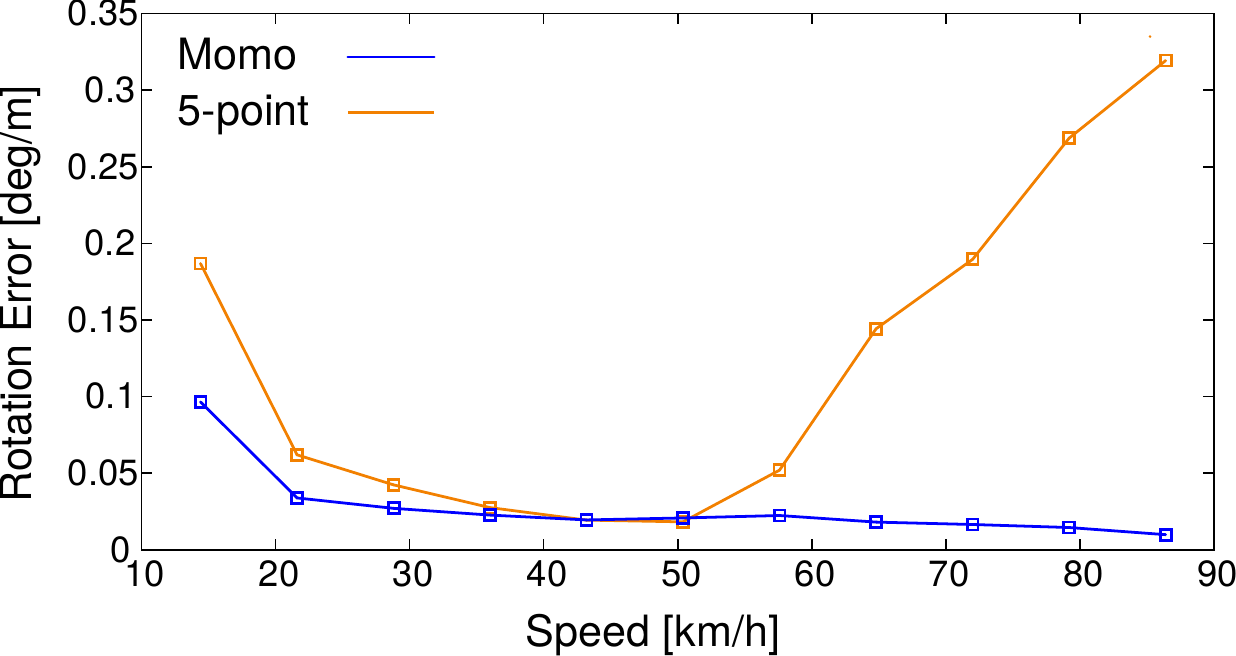}%
\label{fig:rotation_error:speed}}
\caption{Error in rotation over travelled distance and speed from the evaluation on the KITTI dataset for Momo and the 5-point algorithm.
Even though for the 5-point-algorithm 1800--2500 matches per image pair are used and for Momo 100--300, Momo performs considerably better. Especially at high speed, the usage of the motion model stabilizes the method.
}
\label{fig:rotation_error}
\end{figure}

\begin{figure}[h]
\centering
\includegraphics[width=0.95\columnwidth]{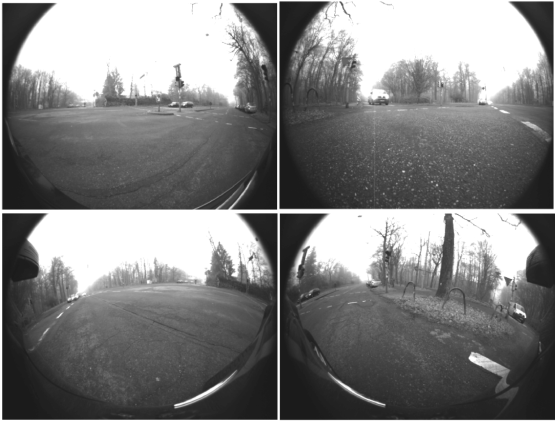}
\caption{Images corresponding to our own multi camera setup.}
\label{fig:camera_images_multi}
\end{figure} 

\begin{figure}[!t]
\centering
\includegraphics[width=0.95\columnwidth]{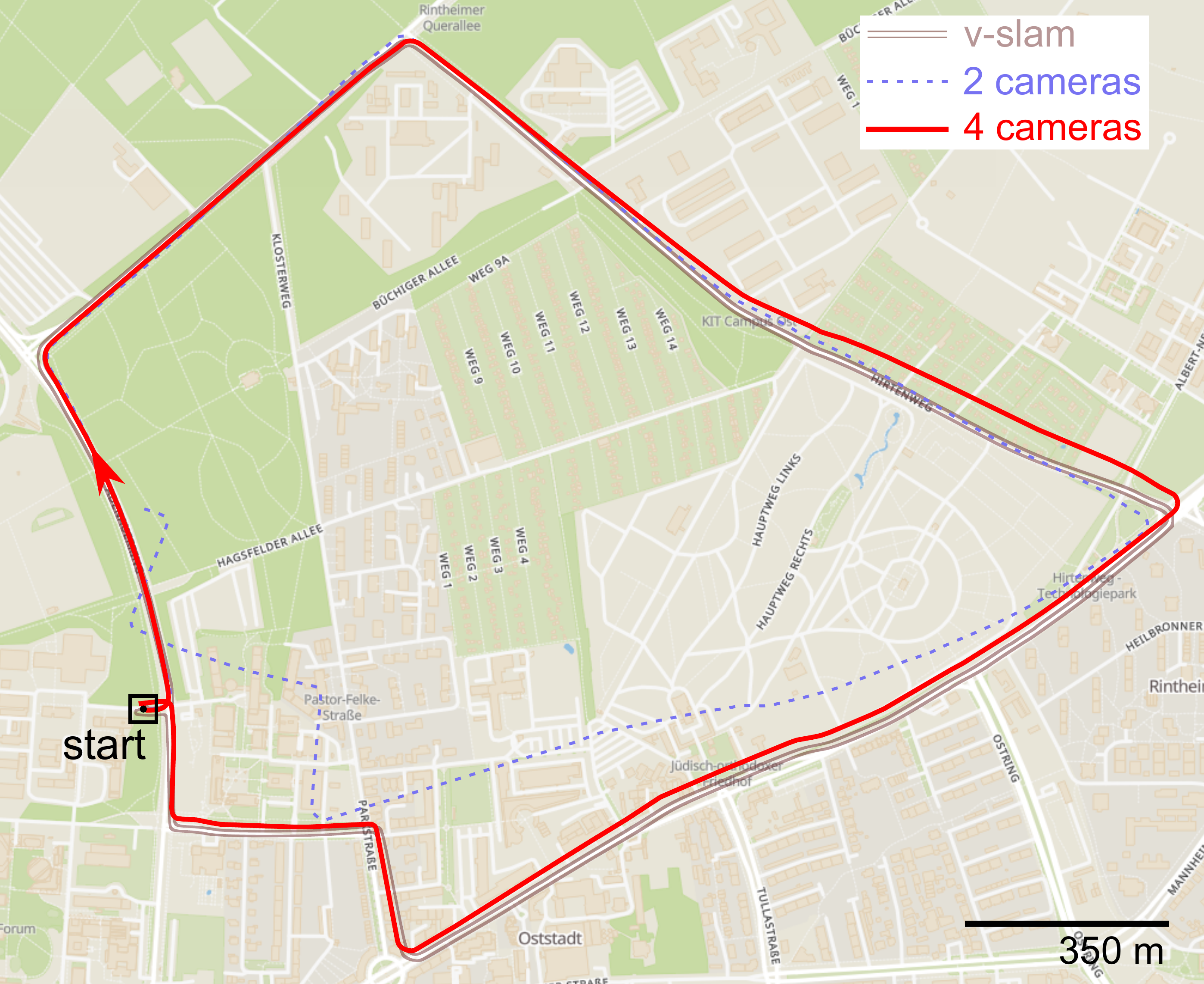}
\caption{Trajectory of the multi camera setup with 4 cameras on a trajectory of $5.1\,\mathrm{km}$ length. Visual SLAM is used as groundtruth (double-line brown). The estimated trajectory of Momo (solid red), operating frame-to-frame, is very close to the groundtruth. The comparison to the trajectory with only the left and the rear camera (dotted blue) shows the benefit of using a surround setup. Especially when the sun blinds the side cameras, the multi camera setup stabilises the estimation substantially. Scale is obtained by the wheel speed of the car.}
\label{fig:eval_bertha_traj}
\end{figure}

\begin{figure}[h]
\centering
\includegraphics[width=0.95\columnwidth]{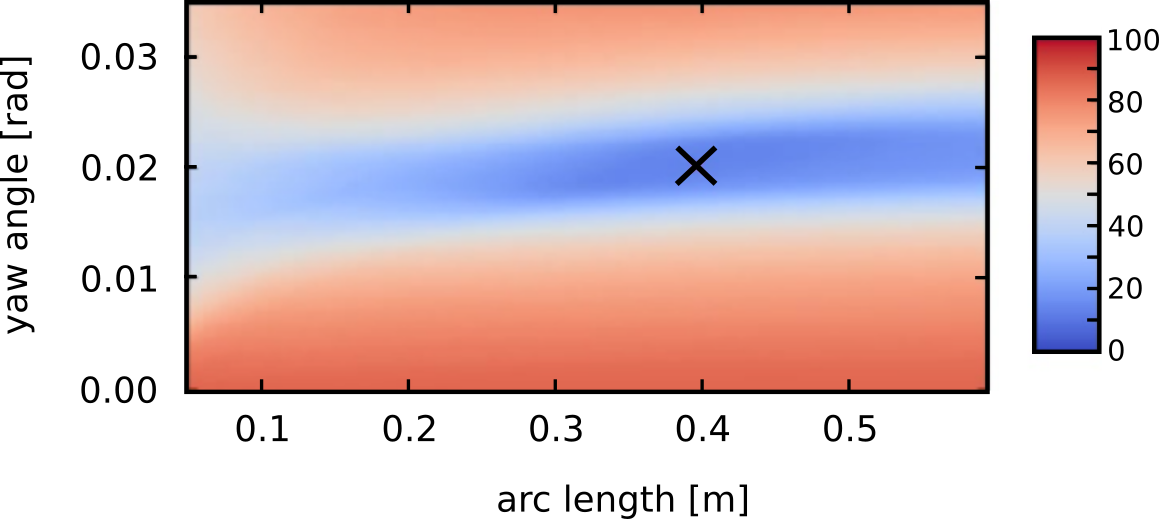}
\caption{Error landscape of the problem in equation~\ref{equ:multiple_cameras} with multiple cameras as shown in fig.~\ref{fig:camera_images_multi} during a curve.
The error is given in percent of maximum error.
The minimum marked by a black cross is observable in both yaw angle and arc length. 
The arc length is observable during the turn since the two side cameras move on circles with different radii.}
\label{fig:error_landscape_scale}
\end{figure} 

\begin{figure}[!t]
\centering
\subfloat[Rotational error over length]{\includegraphics[width=\columnwidth]{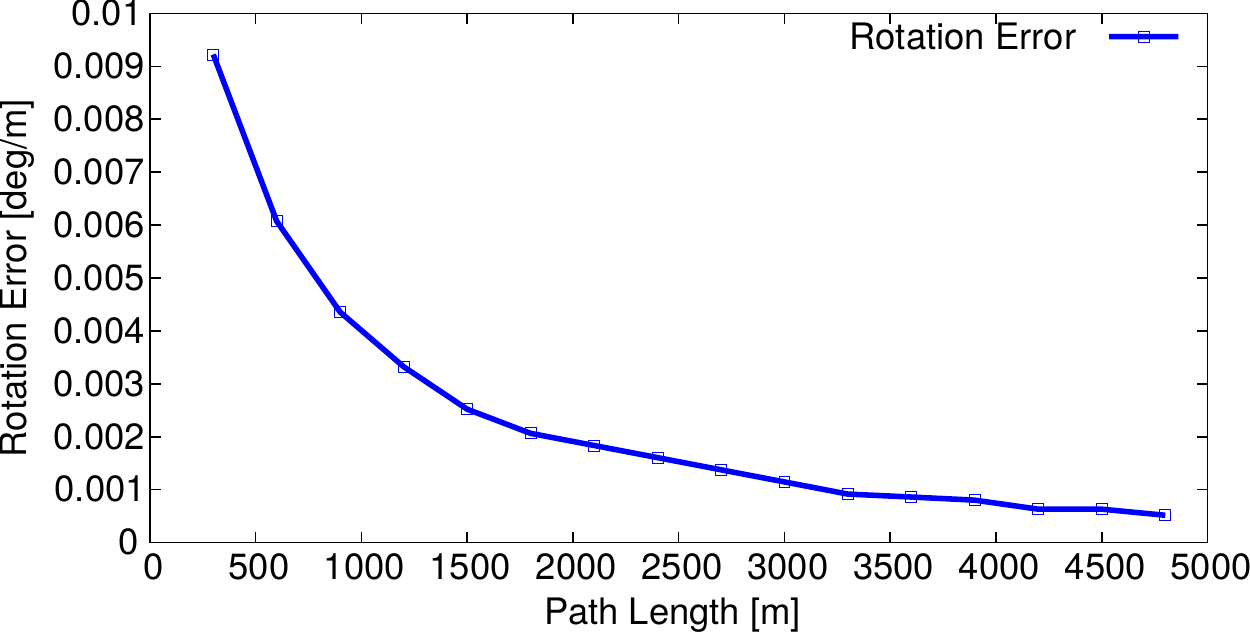}%
\label{fig:eval_bertha_plots:rl}}
\hfil
\subfloat[Rotational error over speed]{\includegraphics[width=\columnwidth]{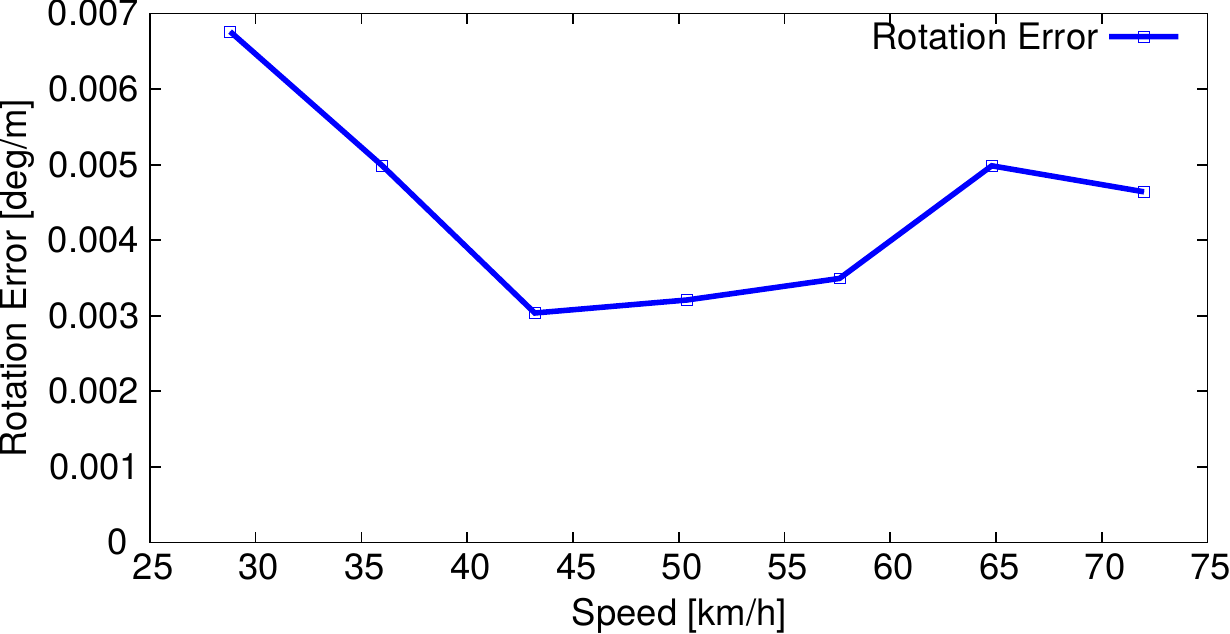}%
\label{fig:eval_bertha_plots:rs}}
\caption{Errors of the multi camera setup shown in fig.~\ref{fig:eval_bertha_traj}, using the error metric of the KITTI-dataset, resulting in rotational error $< 0.001\,\frac{\mathrm{deg}}{\mathrm{m}}$. This is in the league of the top performing stereo and LIDAR methods on the KITTI dataset. }
\label{fig:eval_bertha_plots}
\end{figure}

The proposed methodology was evaluated on the challenging KITTI dataset~\cite{geiger2013vision}. 
The evaluation is effectuated on the public part of the dataset since the groundtruth motion is needed as prior for the arc length.
In order to account for illumination changes the image was gamma corrected.
Subsequently, we used both blobs and corners as key points and the feature descriptor and matching strategy from Geiger et al.~\cite{geiger2011stereoscan} was executed in order to obtain the feature matches $X$. In Momo we use the previously estimated motion as the prior. We set the width of the Cauchy distribution employed in the loss function to $0.0065$. No bundle adjustment was used, only frame-to-frame motion estimation was evaluated.
Two example trajectories from the dataset as well as the average rotational errors over the first 11 sequences are shown in fig.~\ref{fig:kitti_paths} and fig.~\ref{fig:rotation_error}. 
Here we want to show the robustness of our algorithm for rough environments. 
For this objective, the matcher is tuned so that only 100--300 feature matches per image pair are computed, by choosing a large patch size for non-maximum-suppression.
Both, Momo and the 5-point-algorithm with RANSAC of the opencv-library are evaluated. While the 5-point algorithm  is not able to deduce the correct motion from the given set of features, Momo succeeds in correctly estimating a frame-to-frame visual odometry.

\subsection{Evaluation on own dataset}
To show the benefit of using multiple cameras, we evaluated the method on a challenging image sequence in the city of Karlsruhe. 
We used four cameras with viewing angle $110^{\circ}$, images of the setup are shown in fig.~\ref{fig:camera_images_multi}.
Scale was estimated in curves as illustrated in fig.~\ref{fig:error_landscape_scale}. 
For straight movement we employed the odometer of the vehicle.
Since the GNSS pose estimate is not accurate caused by multi-reflections inside the city, we evaluated the accuracy by comparison with a map calculated by classical visual SLAM with loop closure and offline post-processing~(Sons et al.~\cite{sons2015multi}). The results and trajectories are shown in fig.~\ref{fig:eval_bertha_traj} and fig.~\ref{fig:eval_bertha_plots}.\\
Even though this sequence is very challenging since the car drives at $0-72\,\frac{\mathrm{km}}{\mathrm{h}}$ and sun was standing low and blinding the cameras, the estimated trajectory of Momo is very precise, even without using bundle adjustment.
Consequently, the method is usable as visual odometry with estimation runtime between $5\,\mathrm{ms}$ and $20\,\mathrm{ms}$ on a consumer laptop.
\section{Conclusion}
In this work it was shown how dropping the linearisation for prior estimation leads to a more reliable and more robust motion estimation.
Taking into account a motion model into the problem and thus optimizing not on the full six dimensional motion space but on manifolds of this space is the key to reject outliers without the need of randomized sampling and hence obtaining precise frame-to-frame visual odometry. 
In order to enable its use in realistic scenarios, the method is designed so that any number of cameras can be included without the need of overlap.
This redundancy enables our method to tolerate malfunctioning or occluded cameras.
On the KITTI-dataset, it was shown that Momo can cope with a very low number of features of around 200, nevertheless estimating the motion correctly.
Additionally, the method was evaluated on a proper multi-camera dataset of $5.1\,\mathrm{km}$ showing precise and robust results.
This methodology estimates a robust motion prior usable in various SLAM applications as well as for localisation in an offline calculated map. 
A video with example scenes as well as the implementation of Momo in C++ and the dataset can be found on GitHub~(\emph{https://github.com/johannes-graeter/momo.git}).
\\
Due to its modular structure, Momo is the fundament for further improvement.
Since complex motion models can be employed and only a small number of features is needed for a good motion estimation, the next step is to extend the framework to motion estimation of moving objects.

% % can use a bibliography generated by BibTeX as a .bbl file
% % BibTeX documentation can be easily obtained at:
% % http://mirror.ctan.org/biblio/bibtex/contrib/doc/
% % The IEEEtran BibTeX style support page is at:
% % http://www.michaelshell.org/tex/ieeetran/bibtex/
\bibliographystyle{acm}
% % argument is your BibTeX string definitions and bibliography database(s)
\bibliography{ms}

\begin{thebibliography}{10}

\bibitem{bradski2008learning}
{\sc Bradski, G., and Kaehler, A.}
\newblock {\em Learning OpenCV: Computer vision with the OpenCV library}.
\newblock " O'Reilly Media, Inc.", 2008.

\bibitem{buczko2016flow}
{\sc Buczko, M., and Willert, V.}
\newblock Flow-decoupled normalized reprojection error for visual odometry.
\newblock In {\em Intelligent Transportation Systems (ITSC), 2016 IEEE 19th
  International Conference on\/} (2016), IEEE, pp.~1161--1167.

\bibitem{cvivsic2015stereo}
{\sc Cvi{\v{s}}i{\'c}, I., and Petrovi{\'c}, I.}
\newblock Stereo odometry based on careful feature selection and tracking.
\newblock In {\em Mobile Robots (ECMR), 2015 European Conference on\/} (2015),
  IEEE, pp.~1--6.

\bibitem{dellaert2012factor}
{\sc Dellaert, F.}
\newblock Factor graphs and gtsam: A hands-on introduction.
\newblock Tech. rep., Georgia Institute of Technology, 2012.

\bibitem{fischler1981random}
{\sc Fischler, M.~A., and Bolles, R.~C.}
\newblock Random sample consensus: a paradigm for model fitting with
  applications to image analysis and automated cartography.
\newblock {\em Communications of the ACM 24}, 6 (1981), 381--395.

\bibitem{geiger2013vision}
{\sc Geiger, A., Lenz, P., Stiller, C., and Urtasun, R.}
\newblock Vision meets robotics: The kitti dataset.
\newblock {\em The International Journal of Robotics Research 32}, 11 (2013),
  1231--1237.

\bibitem{geiger2011stereoscan}
{\sc Geiger, A., Ziegler, J., and Stiller, C.}
\newblock Stereoscan: Dense 3d reconstruction in real-time.
\newblock In {\em Intelligent Vehicles Symposium (IV), 2011 IEEE\/} (2011),
  Ieee, pp.~963--968.

\bibitem{hartley2003multiple}
{\sc Hartley, R., and Zisserman, A.}
\newblock {\em Multiple view geometry in computer vision}.
\newblock Cambridge university press, 2003.

\bibitem{hartley1997defense}
{\sc Hartley, R.~I.}
\newblock In defense of the eight-point algorithm.
\newblock {\em IEEE Transactions on pattern analysis and machine intelligence
  19}, 6 (1997), 580--593.

\bibitem{hee2013motion}
{\sc Hee~Lee, G., Faundorfer, F., and Pollefeys, M.}
\newblock Motion estimation for self-driving cars with a generalized camera.
\newblock In {\em Proceedings of the IEEE Conference on Computer Vision and
  Pattern Recognition\/} (2013), pp.~2746--2753.

\bibitem{kummerle2011g}
{\sc K{\"u}mmerle, R., Grisetti, G., Strasdat, H., Konolige, K., and Burgard,
  W.}
\newblock g 2 o: A general framework for graph optimization.
\newblock In {\em Robotics and Automation (ICRA), 2011 IEEE International
  Conference on\/} (2011), IEEE, pp.~3607--3613.

\bibitem{nister2004efficient}
{\sc Nist{\'e}r, D.}
\newblock An efficient solution to the five-point relative pose problem.
\newblock {\em IEEE transactions on pattern analysis and machine intelligence
  26}, 6 (2004), 756--770.

\bibitem{rousseeuw2005robust}
{\sc Rousseeuw, P.~J., and Leroy, A.~M.}
\newblock {\em Robust regression and outlier detection}, vol.~589.
\newblock John wiley \& sons, 2005.

\bibitem{scaramuzza2011}
{\sc Scaramuzza, D.}
\newblock 1-point-ransac structure from motion for vehicle-mounted cameras by
  exploiting non-holonomic constraints.
\newblock {\em International journal of computer vision 95}, 1 (2011), 74--85.

\bibitem{sons2015multi}
{\sc Sons, M., Lategahn, H., Keller, C.~G., and Stiller, C.}
\newblock Multi trajectory pose adjustment for life-long mapping.
\newblock In {\em Intelligent Vehicles Symposium (IV), 2015 IEEE\/} (2015),
  IEEE, pp.~901--906.

\bibitem{torr2004invariant}
{\sc Torr, P.~H., and Fitzgibbon, A.~W.}
\newblock Invariant fitting of two view geometry.
\newblock {\em IEEE transactions on pattern analysis and machine intelligence
  26}, 5 (2004), 648--650.

\end{thebibliography}
\newpage

\end{document}